%% file: main.tex
\documentclass[10pt,twocolumn,letterpaper]{article}

\usepackage[pagenumbers]{cvpr} 

\usepackage{graphicx}
\usepackage{amsmath}
\usepackage{amssymb}
\usepackage{booktabs}

\usepackage{mathtools}
\usepackage{amsthm}

\usepackage{xcolor}
\usepackage{multirow}

\usepackage{subcaption}

\usepackage[accsupp]{axessibility}

\usepackage{xcolor, soul}
\sethlcolor{pink}

\usepackage[pagebackref,breaklinks,colorlinks]{hyperref}

\usepackage[capitalize]{cleveref}
\crefname{section}{Sec.}{Secs.}
\Crefname{section}{Section}{Sections}
\Crefname{table}{Table}{Tables}
\crefname{table}{Tab.}{Tabs.}

\newcommand{\diff}{\mathop{}\!\mathrm{d}}
\newcommand{\expp}{\mathrm{e}}
\newcommand{\cond}{{\;|\;}}

\def\cvprPaperID{86} 
\def\confName{CVPR}
\def\confYear{2023}

\begin{document}


\title{Refusion: Enabling Large-Size Realistic Image Restoration with \\ Latent-Space Diffusion Models}

\author{Ziwei Luo \, Fredrik K. Gustafsson \, Zheng Zhao \, Jens Sjölund \, Thomas B. Schön\\
Uppsala University, Sweden\\
{\tt\small \{ziwei.luo,fredrik.gustafsson,zheng.zhao,jens.sjolund,thomas.schon\}@it.uu.se} \\
{\tt\small\url{https://github.com/Algolzw/image-restoration-sde}}
}

\maketitle

\begin{abstract}
   \input{sections/abstract}
\end{abstract}

\input{sections/introduction}

\input{sections/related_work}

\input{sections/method}

\input{sections/experiment}
\input{sections/conclusion}

\vspace{5pt}
\noindent \textbf{Acknowledgements}
This research was partially supported by the \emph{Wallenberg AI, Autonomous Systems and Software Program (WASP)} funded by Knut and Alice Wallenberg Foundation, by the project \emph{Deep probabilistic regression -- new models and learning algorithms} (contract number: 2021-04301) funded by the Swedish Research Council, and by the
\emph{Kjell \& M{\"a}rta Beijer Foundation}.
The computations were enabled by the \textit{Berzelius} resource provided by the Knut and Alice Wallenberg Foundation at the National Supercomputer Centre.



{\small
\bibliographystyle{ieee_fullname}
\bibliography{main}
}

\end{document}

%% file: sections/abstract.tex
This work aims to improve the applicability of diffusion models in realistic image restoration. Specifically, we enhance the diffusion model in several aspects such as network architecture, noise level, denoising steps, training image size, and optimizer/scheduler. We show that tuning these hyperparameters allows us to achieve better performance on both distortion and perceptual scores. We also  propose a U-Net based latent diffusion model which performs diffusion in a low-resolution latent space while preserving high-resolution information from the original input for the decoding process. Compared to the previous latent-diffusion model which trains a VAE-GAN to compress the image, our proposed U-Net compression strategy is significantly more stable and can recover highly accurate images without relying on adversarial optimization. Importantly, these modifications allow us to apply diffusion models to various image restoration tasks, including real-world shadow removal, HR non-homogeneous dehazing, stereo super-resolution, and bokeh effect transformation. By simply replacing the datasets and slightly changing the noise network, our model, named \textbf{Refusion}, is able to deal with large-size images (e.g., $6000 \times 4000 \times 3$ in HR dehazing) and produces good results on all the above restoration problems.
Our Refusion achieves the best perceptual performance in the NTIRE 2023 Image Shadow Removal Challenge and wins 2$^{\text{nd}}$ place overall. 

%% file: sections/introduction.tex
\section{Introduction}

Image restoration is a long-standing problem in computer vision due to its ill-posed nature and extensive demands in industry. Broadly speaking the challenge is to restore the high-quality (HQ) image from the low-quality (LQ) counterpart subject to various degradation factors (e.g., noising, downsampling, and hazing). Over the past decade, methods based on deep learning have achieved impressive performance in image restoration. However, most of these methods are prone to produce over-smooth images due to their pixel-based reconstruction loss functions, i.e., $L_1/L_2$~\cite{dong2015image,zhang2017beyond,tao2018scale,ren2016single,ren2019progressive}.

Recently, the diffusion model has shown a strong capability in producing high-quality results by sampling images consisting of pure noise and then iteratively denoising them with Langevin dynamics~\cite{ho2020denoising,song2020denoising,dhariwal2021diffusion} or reverse-time stochastic differential equations (SDEs)~\cite{song2020score,vahdat2021score}. However, many common image restoration tasks (e.g., deraining, dehazing, and deblurring) are still challenging for diffusion models, due to the complex degradations and the large and arbitrary image sizes in real-world datasets. 
There has been interesting developments when it comes to the use of pre-trained diffusion models. Two drawbacks with the existing approaches are that; 1) they rely on carefully curating the datasets (e.g., ImageNet~\cite{deng2009imagenet} and FFHQ~\cite{karras2019style}), 2) they require the degradation parameters to be known.
These drawbacks limit their applicability when it comes to real-world  tasks~\cite{kawar2021snips,kawar2022denoising,bansal2022cold,chung2022improving,daras2022soft,choi2021ilvr}. 

In order to handle intricate real-world distortions, recent developments~\cite{ozdenizci2023restoring,whang2022deblurring,saharia2022image} have utilized a combination of a pure noise image and a low-quality image as an intermediary input for the noise network. This approach avoids the need for degradation parameters and enforces the reverse process to convert the noise into the desired high-quality image. However, these approaches are somewhat heuristic and are difficult to apply to general tasks. A more general image restoration method is IR-SDE~\cite{luo2023image}, which proposes to recover HQ images based on a mean-reverting SDE, which implicitly models the degradation and is applicable to various tasks by changing the datasets only. A drawback with IR-SDE is that it is computationally demanding at test time since it requires multi-step denoising on the full image to restore the final output. This can be problematic for real-world applications, in particular for high-resolution images.

The purpose of this paper is to improve the diffusion model is a way that enhance its effectiveness in tackling diverse real-world image restoration tasks. The result is \textbf{Refusion} (\textit{image \textbf{Re}storation with dif\textbf{fusion} models}). 
Due to its simplicity and flexibility in accommodating different problems, the IR-SDE serves as the foundation for Refusion. By exploring different noise network architectures, we show that using the nonlinear-activation-free network (NAFNet)~\cite{chen2022simple} can achieve good performance in noise/score prediction while at the same time being more computationally efficient. Moreover, we also illustrate the efficacy of different noise levels, denoising steps, training image sizes, and optimizer/scheduler selections. To further deal with large images, we propose a U-Net based latent diffusion strategy. This allows us to perform image restoration in a compressed and low-resolution latent space, which speeds up both the training and the inference. 
In the experiments, we demonstrate our improved diffusion model on the tasks of real-world shadow removal, HR non-homogeneous dehazing (with images of size $6000 \times 4000 \times 3$), stereo super-resolution, and the bokeh effect transformation. The experiments show that the proposed Refusion is effective on all the image restoration tasks mentioned above.

Our contributions are summarized as follows:
\begin{itemize}

    \item Compared to existing diffusion-based approaches, our method can gracefully handle high-resolution images by performing image restoration in the U-Net compressed latent space, while preserving high-resolution information from the original input for the decoding process. Importantly, our U-Net compression strategy offers significantly improved stability compared to existing latent diffusion models and can recover high-accuracy images without requiring adversarial optimization.

    \item We perform a comprehensive empirical study of several factors that have a major impact on the performance of diffusion models for image restoration.

    \item We propose to change the diffusion base network from U-Net to NAFNet. The latter achieves better image restoration performance across all tasks while requiring fewer model parameters and being computationally more efficient.


    \item We evaluate our approach on extensive real-world and synthetic datasets, further showing strong versatility to a variety of image restoration problems.
\end{itemize}

%% file: sections/related_work.tex
\begin{figure*}[t]
\begin{center}
\includegraphics[width=1.0\linewidth]{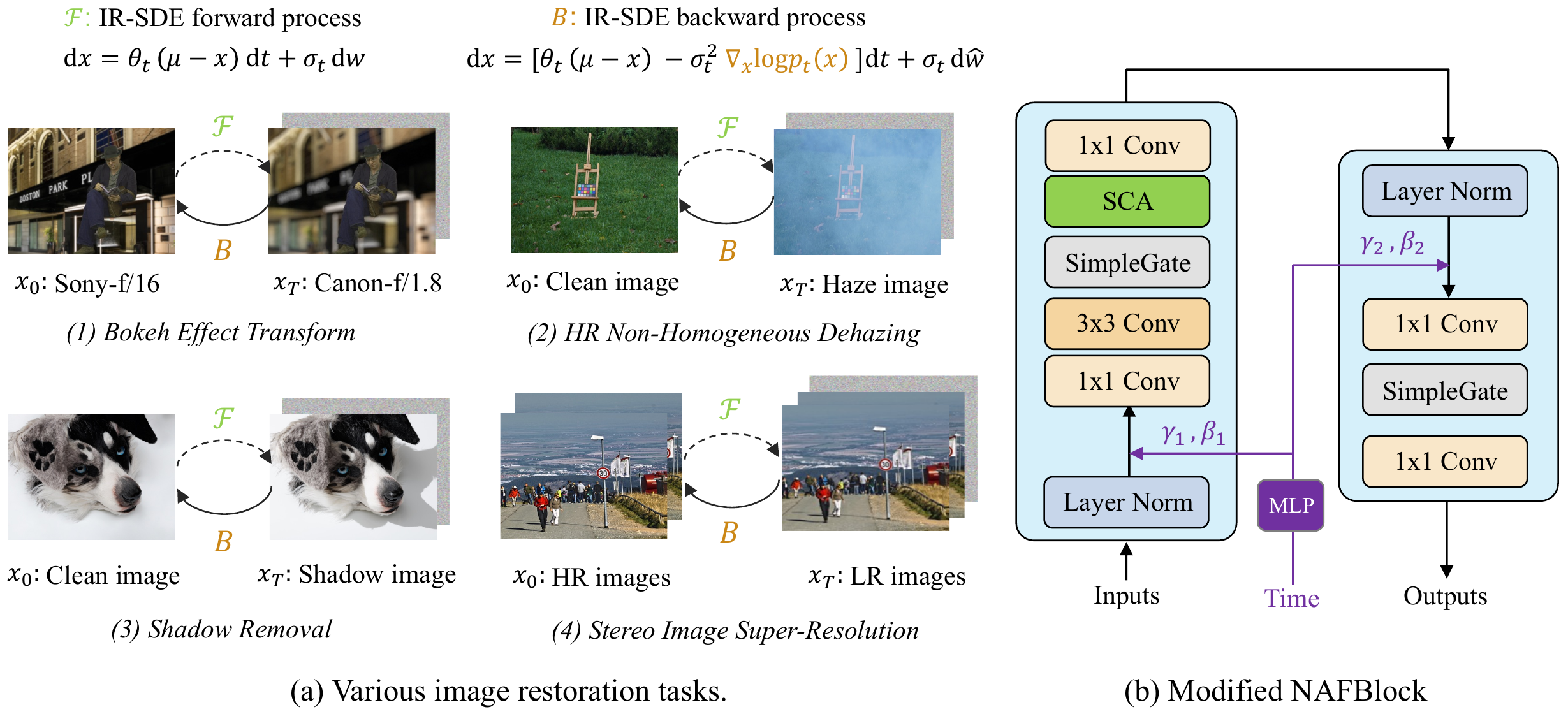}\vspace{-4.0mm}
\end{center}
    \caption{(a) Illustrations of various image restoration tasks based on our proposed Refusion method. We use the mean-reverting SDE to recover HQ images from LQ images, adopting the IR-SDE approach~\cite{luo2023image}. (b) The NAFBlock with an additional time processing branch which is depicted in purple color. Here ``SCA'' is the simple channel attention, and ``SimpleGate'' is an element-wise operation that splits feature channels into two parts and then multiplies them as output.}
\label{fig:overview}
\end{figure*}

\section{Related Work}

Image restoration aims to restore a high-quality image from a degraded low-quality version. When it comes to approaches based on deep learning, two early and influential contributions are SRCNN~\cite{dong2015image} and DnCNN~\cite{zhang2017beyond}. They made use of convolution neural networks (CNNs) for image super-resolution and denoising, which significantly improved the performance in each application. This development spurred a lot of activity when it comes to making use of CNNs for various image restoration tasks~\cite{dong2016accelerating,zhang2018residual,kim2016accurate,li2022d2c,luo2022deep,zamir2021multi,zhang2017learning,zamir2020learning,lian2023kernel,zhang2021plug,wang2018esrgan,cheng2021nbnet,zhang2020residual,luo2023fast,chen2021hinet}. Many of these approaches can be viewed as variations of~\cite{dong2015image, zhang2017beyond} trained with pixel reconstruction losses such as $L_1$ and  $GAN$. 

Recently, transformer-based architectures~\cite{vaswani2017attention} have shown impressive performance and hence received a lot of attention when it comes to high-level computer vision tasks~\cite{dosovitskiy2020image,liu2021swin,he2022masked}. These architectures have also been employed for image restoration~\cite{chen2021pre,liang2021swinir,zamir2022restormer,wang2022uformer,luo2022bsrt,xiao2022image}. For example, IPT~\cite{chen2021pre} is the first work to propose the use of pre-trained transformers for image processing. Subsequently, SwinIR~\cite{liang2021swinir} modifies the Swin Transformer~\cite{liu2021swin} with additional convolution layers and residual connections to achieve state-of-the-art performance on various image restoration tasks such as image super-resolution and denoising. Restormer~\cite{zamir2022restormer} and Uformer~\cite{wang2022uformer} combine the transformer with U-shape structures to achieve more efficient image restoration. In addition, there are also attempts to make use of the MLP~\cite{tu2022maxim} and the nonlinear activation free networks~\cite{chen2022simple} to restore images.

%% file: sections/method.tex
\section{Preliminaries: Mean-Reverting SDE}
\label{section:preliminaries}

Our method leverages a diffusion model for realistic image restoration. Specifically, we use IR-SDE~\cite{luo2023image} as the base diffusion framework, which can naturally transform the high-quality image to its degraded low-quality counterpart, irrespective of how complicated the degradation is (even for real-world degradations, see~\Cref{fig:overview}). The forward process of the IR-SDE is defined as:
\begin{equation}
	dx = \theta_t \, (\mu - x) d t + \sigma_t d w, 
	\label{equ:ou}
\end{equation}
where $\theta_t$ and $\sigma_t$ are time-dependent positive parameters characterizing the mean-reversion speed and the stochastic volatility, respectively. If we set the SDE coefficients in~\eqref{equ:ou} to satisfy $\sigma_t^2 \, / \, \theta_t = 2 \, \lambda^2$ for all times $t$, the marginal distribution $p_t(x)$ can be computed according to~\cite{luo2023image}
\begin{subequations}
\begin{align}
    p_t(x) & \ = \mathcal{N}\bigl(x(t) \cond m_t, v_{t}\bigr),\\
    m_t &\coloneqq \mu + (x(0) - \mu) \, \expp^{-\bar{\theta}_{t}}, \\
    v_{t} & \coloneqq \lambda^2 \, \Bigl(1 - \expp^{-2 \, \bar{\theta}_{t}}\Bigr),
\end{align}\label{eq:sde_solution}
\end{subequations}
where $\bar{\theta}_{t} \coloneqq \int^t_0 \theta_z \diff z$. Note that as~$t$ increases, the mean value $m_t$ and the variance $v_{t}$ converges to $\mu$ and $\lambda^2$, respectively. Hence, the initial state~$x(0)$ is iteratively transformed into~$\mu$ with additional noise, where the noise level is fixed to~$\lambda$.

The IR-SDE forward process (\ref{equ:ou}) is a forward-time It\^{o} SDE, which has a reverse-time representation as~\cite{song2020score}
\begin{equation}
    dx = \big[ \theta_t \, (\mu - x) - \sigma_t^2 \, \nabla_{x} \log p_t(x) \big] d t + \sigma_t d \hat{w}.
    \label{eq:reverse-irsde}
\end{equation}
Note that during training we have access to HQ images which means that we can employ~\eqref{eq:sde_solution} to compute the ground truth score function
\begin{equation}
    \nabla_{x}\log p_t(x) = - \frac{x(t) - m_t}{v_t}.
    \label{eq:score}
\end{equation}
The reparameterization trick now allows us to sample $x(t)$ according to $x(t) = m_{t}(x) + \sqrt{v_{t}} \, \epsilon_t$, where $\epsilon_t \sim \mathcal{N}(0, I)$ is a standard Gaussian noise. Then we can rewrite \eqref{eq:score} as $\nabla_{x}\log p_t(x) = - \frac{\epsilon_t}{v_t}$. A CNN network is usually trained to estimate the noise, and at test time we then simulate the backward SDE to transform low-quality images into high-quality versions, similar to other diffusion-based models.

\section{Improving the Diffusion Model}
\label{sec:method}

We will in~\cref{subsection:latent_diffusion} introduce the U-Net based latent diffusion model that allows us to perform diffusion in the low-resolution space to significantly improve the sample efficiency. After that, we introduce the nonlinear activation free blocks (NAFBlocks)~\cite{chen2022simple} to IR-SDE in~\cref{subsection:naf}, and outline several training strategies that can improve the restoration performance in~\cref{subsection:training}. As an overview, \Cref{fig:overview} illustrates the tasks and networks of the proposed Refusion method.

\begin{figure}[t]
\begin{center}
\includegraphics[width=1.\linewidth]{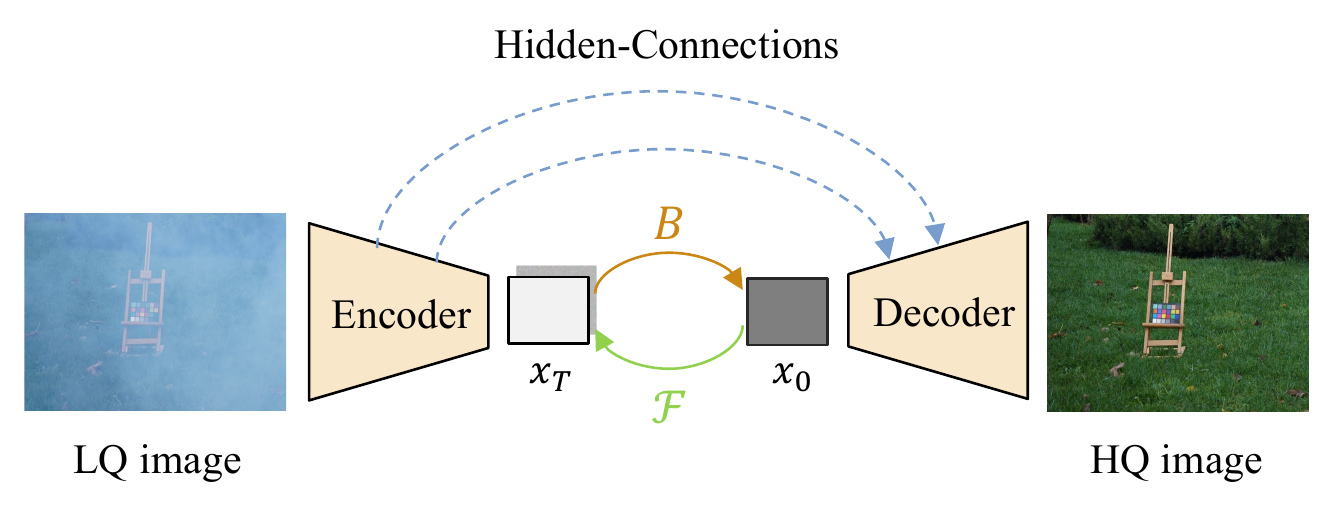}\vspace{-3.0mm}
\end{center}
    \caption{Overview of our U-Net based latent diffusion model. The restoration is performed in a low-resolution latent space.}
\label{fig:latent}
\end{figure} 

\subsection{Latent Diffusion under U-Net}
\label{subsection:latent_diffusion}

Iteratively running the diffusion model (even with just a few denoising steps) on tasks with high-resolution images is notoriously time-consuming. Especially for HR dehazing, where all images are captured with $6000\times 4000 \times 3$ pixels, which is far beyond the input size of any existing diffusion model. To handle large input sizes we propose to perform the restoration in a low-resolution latent space, by incorporating a pretrained U-Net network. The overall architecture of the proposed U-Net based latent diffusion model is shown in~\Cref{fig:latent}. An encoder compresses the LQ image into a latent representation, which is transformed into an HQ latent representation via the IR-SDE backward process. From this, a decoder then reconstructs an HQ image. An important difference compared to latent-diffusion~\cite{rombach2022high}, which uses VAE-GAN as the compressing model, is that the proposed U-Net maintains multi-scale details flowing from the encoder to the decoder through skip-connections. This better captures the input image's information and provides the decoder with additional details to reconstruct more accurate HQ images.

\begin{figure}[t]
\begin{center}
\includegraphics[width=1.\linewidth]{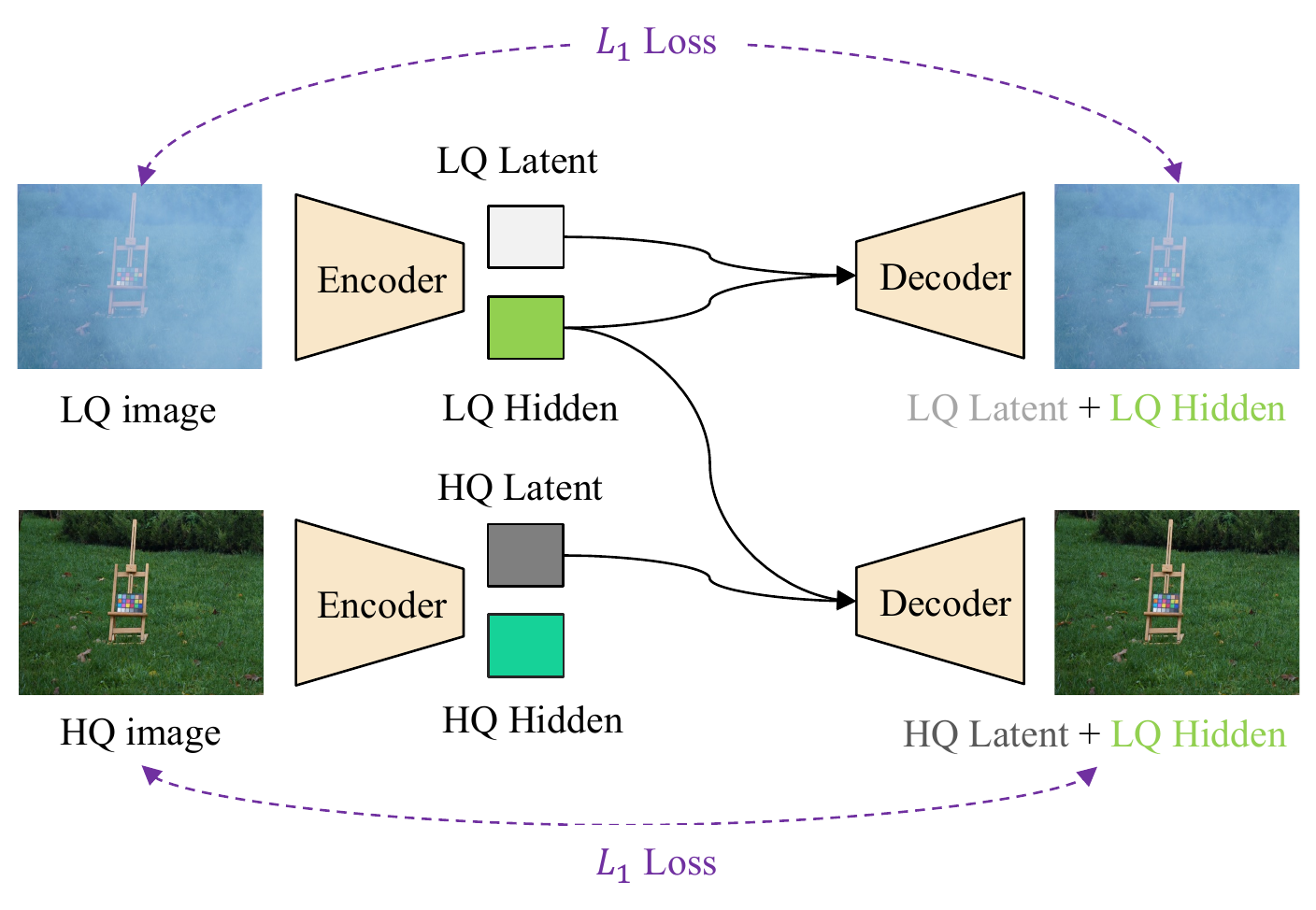}\vspace{-4.0mm}
\end{center}
    \caption{Our proposed latent-replacing pretraining strategy for the U-Net model, utilizing two reconstruction $L_1$ loss terms.}
\label{fig:unet}
\end{figure}

When training the U-Net model, we need to make sure that the compressed latent representation is discriminative and contains the main degradation information. The U-Net decoder must also be able to reconstruct HQ images from transformed LQ latent representations. We therefore adopt a latent-replacing training strategy, as shown in~\Cref{fig:unet}. Each LQ image is first encoded and decoded by the U-Net, and a reconstruction $L_1$ loss is applied. The U-Net is then also trained to reconstruct the corresponding HQ image, by replacing the LQ latent representation with that of the HQ image and running the decoder again. Importantly, our proposed training strategy does not involve any adversarial optimization. The model training is thus more stable than for latent-diffusion~\cite{rombach2022high}.

\begin{figure*}[ht]
\centering
	\begin{minipage}{0.247\linewidth}
		\centering
        \subcaptionbox{\label{subfig:curve_1}}
		{\includegraphics[width=1.\linewidth]{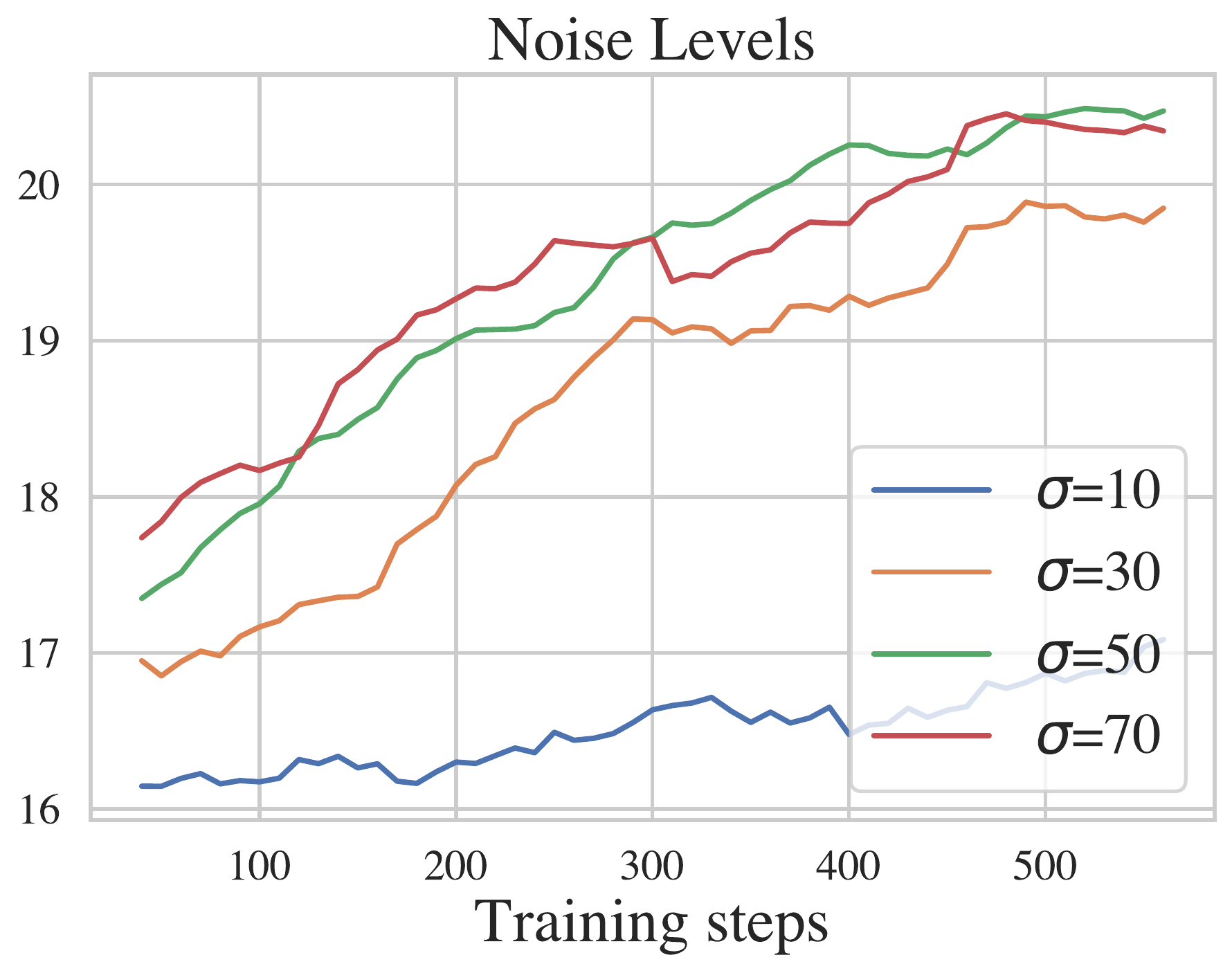}}
    
	\end{minipage}
	\begin{minipage}{0.246\linewidth}
		\centering
        \subcaptionbox{\label{subfig:curve_2}}
		{\includegraphics[width=1.\linewidth]{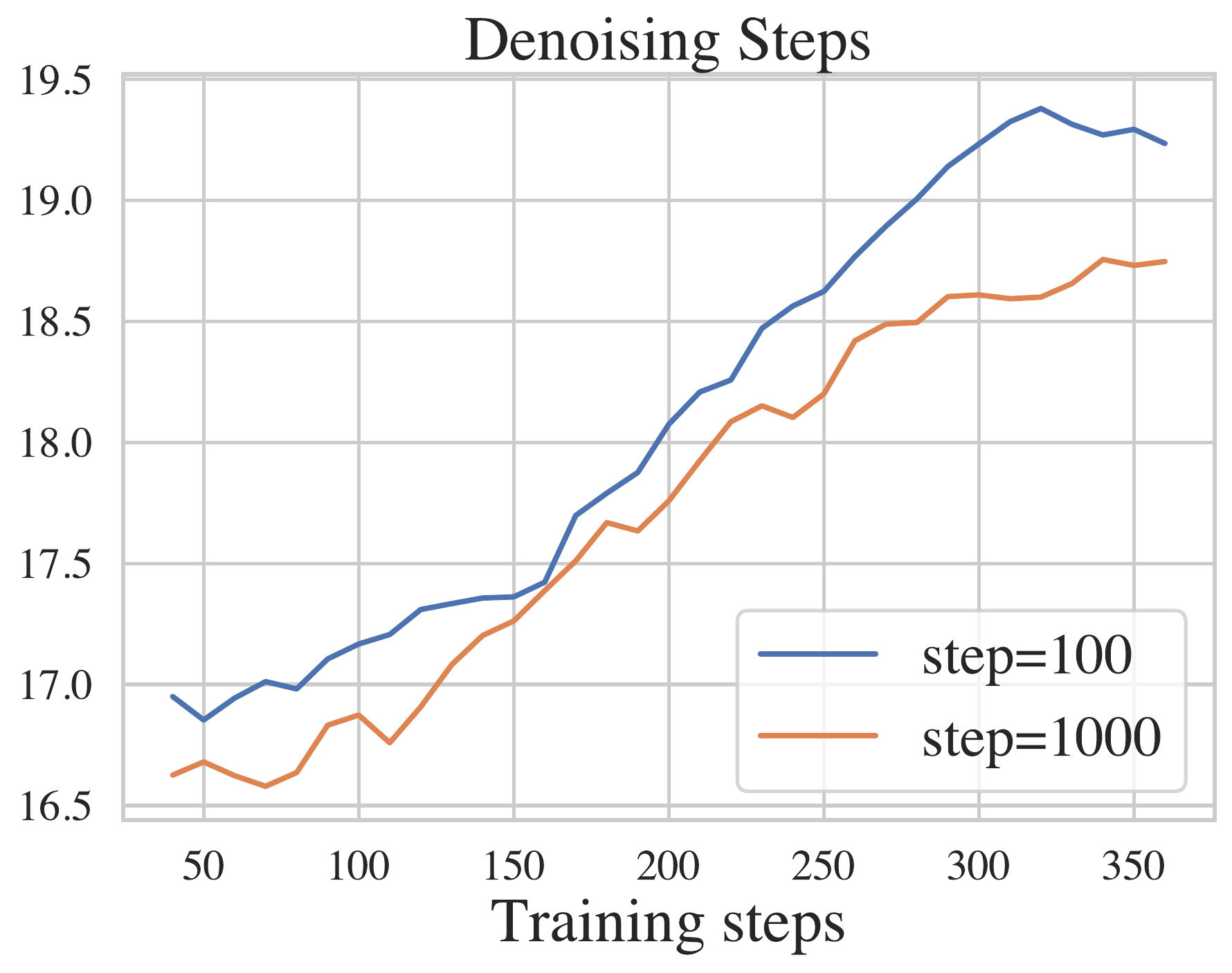}}
	\end{minipage}
	\begin{minipage}{0.246\linewidth}
		\centering
        \subcaptionbox{\label{subfig:curve_3}}
		{\includegraphics[width=1.\linewidth]{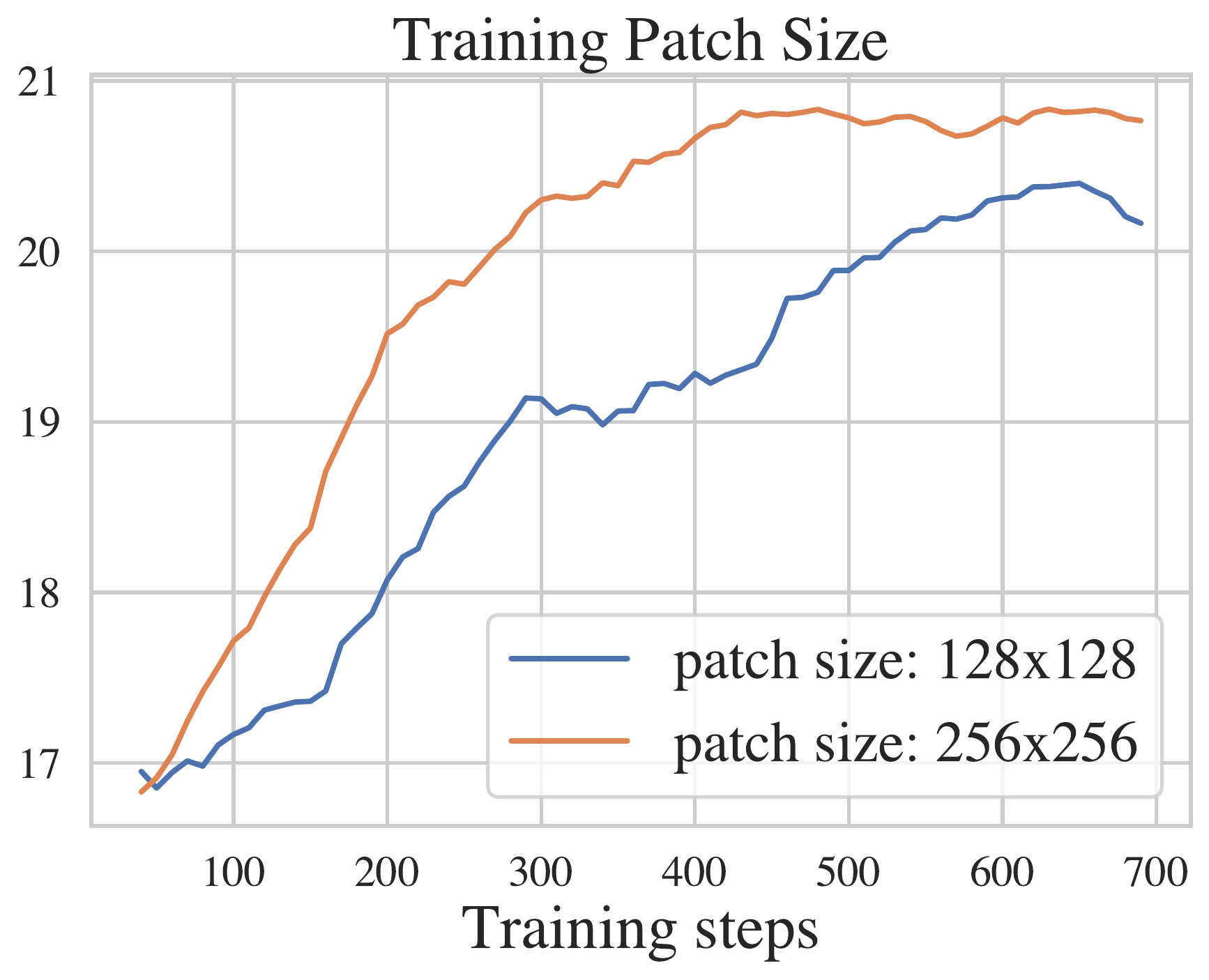}}
	\end{minipage}
    \begin{minipage}{0.246\linewidth}
		\centering
        \subcaptionbox{\label{subfig:curve_4}}
		{\includegraphics[width=1.\linewidth]{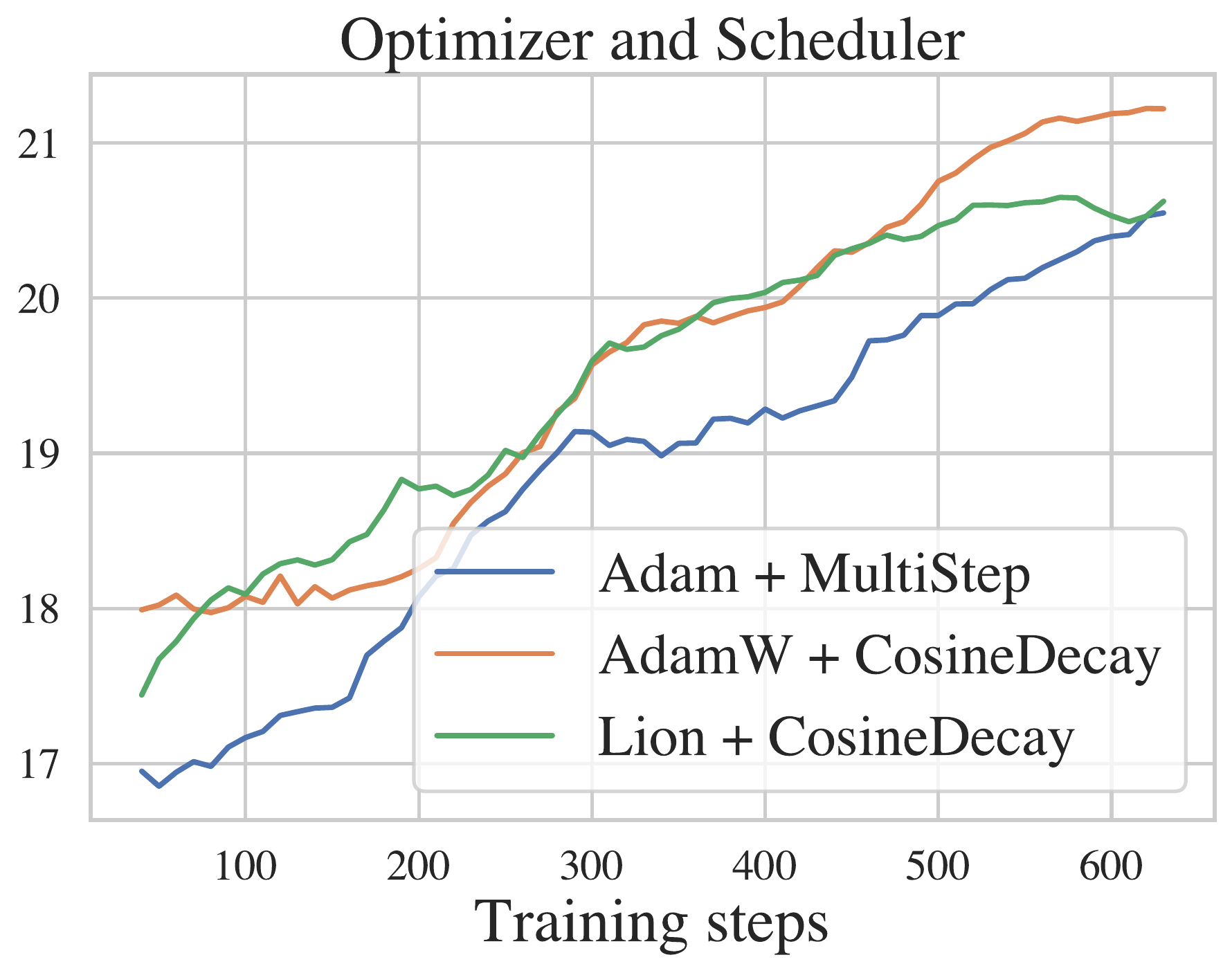}}
	\end{minipage}\vspace{-2.0mm}
	\caption{Validation PSNR curves of different training strategies on the real-world shadow removal task. All models use the same modified NAFNet backbone. We show that slightly changing these parameters can lead to significant performance improvements.}
	\label{fig:curves}
\end{figure*}

\subsection{Modified NAFBlocks for Noise Prediction}
\label{subsection:naf}

A commonly used architecture for noise/score prediction is the U-Net~\cite{ronneberger2015u} with residual blocks~\cite{he2016deep} and attention mechanisms such as channel-attention and self-attention~\cite{ho2020denoising,song2020score}. 
The recently proposed DiT~\cite{peebles2022scalable} makes use of a transformer-based structure and simulates diffusion in a lower-resolution latent space, which sets a new state-of-the-art on the class-conditional ImageNet $512 \times 512$ generation in terms of FID. But even under the latent-diffusion framework~\cite{rombach2022high}, pure transformer architectures still incur a larger computational cost than affordable in traditional image restoration applications.

\begin{figure}[ht]
\centering
\includegraphics[width=.7\linewidth]{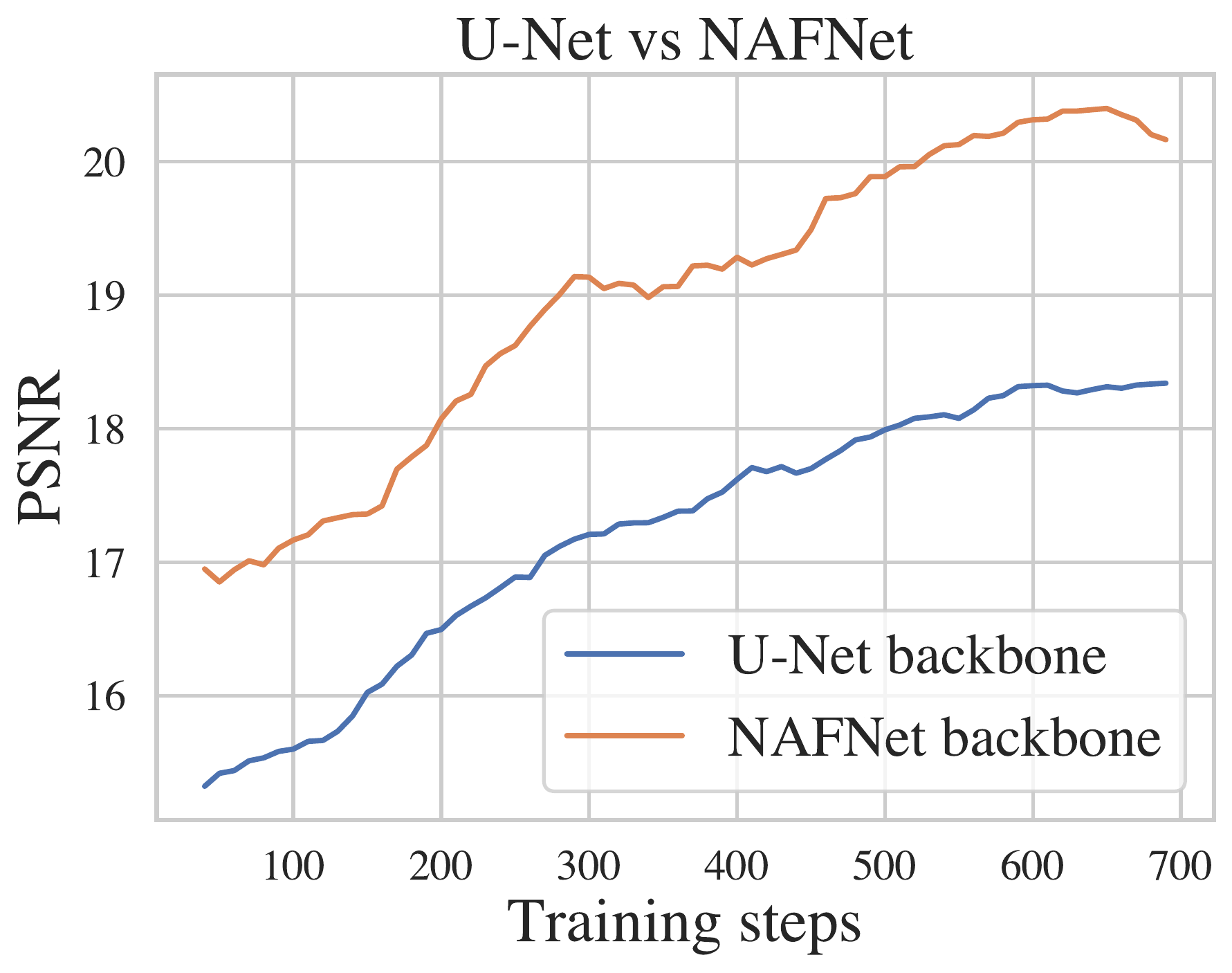}
\caption{Comparison of learning curves between U-Net and the modified NAFNet backbone on the shadow removal dataset.}
\label{fig:unet_nafnet}
\end{figure}

To address the aforementioned model efficiency problem, we explore a new architecture for noise prediction. Specifically, our noise network is based on slightly modified nonlinear activation free blocks (NAFBlocks)~\cite{chen2022simple}. Nonlinear activation free means that we replace all nonlinear activation functions with the ``SimpleGate'', an element-wise operation that splits feature channels into two parts and then multiplies them together to produce the output. As illustrated in~\Cref{fig:overview}(b), we add an additional multilayer perceptron to process the time embedding to channel-wise scale and shift parameters $\gamma$ and $\beta$, for both the attention layer and feed-forward layer. To adapt to different tasks, we also slightly modify the network with task-specific architectures, such as the lens information in \textit{Bokeh Effect Transform} and dual inputs in \textit{Stereo Image Super-Resolution}. 
The learning curves of U-Net and NAFNet based diffusion are illustrated in~\Cref{fig:unet_nafnet}. Note that the modified NAFNet significantly outperforms the U-Net backbone on the shadow removal task.

\subsection{Improved Training Strategies}
\label{subsection:training}
In this section, we discuss the main factors that affect the training process of diffusion-based restoration. The analysis is  performed using the real-world shadow removal task.

\vspace{5pt}
\noindent \textbf{Noise levels.}
The noise level (i.e., the stationary variance~$\lambda$ from Section~\ref{section:preliminaries}) can play an important role when it comes to the performance of diffusion models~\cite{nichol2021improved,luo2023image}. 
In image restoration we often recover an HQ image directly from the LQ image rather than from pure noise, which means that a standard Gaussian for the terminal state is not necessary.  
As shown in~\Cref{subfig:curve_1}, we compare four different noise levels $\sigma=\{10, 30, 50, 70\}$ on the shadow removal task. The training curves show that setting~$\sigma=50$ or~$\sigma=70$ is more stable than that with a small noise level, i.e. $\sigma=10$.

\vspace{5pt}
\noindent \textbf{Denoising steps.}
Several works propose to use long-step pretrained weights but generating images using fewer steps~\cite{song2020denoising,lu2022dpm,bao2022analytic}, which indeed improves the sample efficiency but however also at the cost of decreasing the image quality. 
In image restoration, we have to re-train diffusion models from scratch for all tasks. 
Since IR-SDE has a stable and robust learning process, we consider to directly adjust the denoising steps in training while maintaining the performance. \Cref{subfig:curve_2} compares the training curves of IR-SDE with 100 and 1000 denoising steps. We find that using fewer denoising steps can result in comparable---and sometimes even better---restoration performance.

\vspace{5pt}
\noindent \textbf{Training patch sizes.}
A common practice is that training with large patches can improve the image restoration performance~\cite{liang2021swinir,lin2022revisiting}. But none of the existing works discussed the effect of patch sizes in training diffusion models. Here we present the comparison of training IR-SDE with patch size $128 \times 128$ and $256 \times 256$, as shown in~\Cref{subfig:curve_3}. As can be observed, training with large patches performs much better, which is consistent with other CNN/Transformer based image restoration approaches.

\vspace{5pt}
\noindent \textbf{Optimizer/scheduler.}
A good optimizer with a proper learning rate scheduler is also important to the performance. As an example, simply adding a cosine decay scheduler can improve the accuracy by 0.5\% for ResNet-50 on the ImageNet classification task~\cite{he2019bag}. 
To find out which optimizer better matches the diffusion model, we provide three comparisons including 1) Adam + multi-step decay, 2) AdamW~\cite{loshchilov2017decoupled} + cosine decay, 3) Lion~\cite{chen2023symbolic} + cosine decay. The results in~\Cref{subfig:curve_4} show that both AdamW and Lion perform slightly better then the Adam optimizer with multi-step learning rate decay.


%% file: sections/experiment.tex
\section{Experiments}

We evaluate Refusion on various image restoration tasks. In this section, we first briefly introduce several restoration tasks and their datasets, and then show the comparisons and results of our proposed method with other baselines. Our method achieves the best perceptual performance in the NTIRE 2023 Image Shadow Removal Challenge~\cite{vasseiz2023ntireshr} and wins 2$^{\text{nd}}$ place in terms of overall performance.

\subsection{Tasks and Datasets}


\noindent \textit{Image Shadow Removal} is the task of mapping shadow regions of an image to their shadow-free counterparts, which can enhance the image quality and benefit downstream computer vision tasks~\cite{nadimi2004physical,sanin2010improved,zhang2018improving}. For the dataset, we follow the instructions in the NTIRE 2023 Shadow Removal Challenge~\cite{vasseiz2023ntireshr,vasluianu2023shadow} to use 1\thinspace000 pairs of shadow and shadow-free images for training and 100 shadow images for validation. 

\vspace{5pt}
\noindent \textit{Stereo Image Super-Resolution} is a problem stemming from the growing popularity of dual cameras in modern mobile phones, and aims to recover high-quality images from paired low-quality left and right images with stereo correspondences~\cite{wang2021symmetric,ying2020stereo}. To train and evaluate our model, we use the dataset provided by the NTIRE 2023 Stereo Image Super-Resolution Challenge~\cite{Wang2023NTIRE}, which consists of 800 training stereo images and 112 validation stereo images from the Flickr1024~\cite{wang2019flickr1024} dataset. All low-resolution images are generated by bicubic downsampling.

\vspace{5pt}
\noindent \textit{Bokeh Effect Transformation} is an important task to computational photography, aiming to convert the image's Bokeh effect from the source lens to that of a target lens without harming the sharp foreground regions in the image~\cite{ignatov2020rendering,ignatov2020aim}. For this purpose, we consider the new dataset proposed in NTIRE 2023 Bokeh Effect Transformation challenge~\cite{conde2023ntire_bokeh}, in which 10\thinspace000 pairs of synthetic images with different lens information are used for training, and 500 images with source and target lens information are used for validation. 

\vspace{5pt}
\noindent \textit{HR Non-Homogeneous Dehazing} aims to perform defogging on extremely high-resolution images with heavy non-homogeneous fog, which is challenging to current dehazing approaches. Here we use a new dataset proposed in the NTIRE 2023 HR NonHomogeneous Dehazing competition~\cite{ancuti2023ntire}. This dataset has a large diversity of contents and is collected similar to NH-HAZE~\cite{ancuti2020nh,ancuti2020ntire}, but with all images having $6000 \times 4000 \times 3$ pixels. In addition, it only contains 40 hazy/haze-free image pairs for training and 10 images for validation and testing. Since the clean images of the validation and test datasets are not accessible, we choose to only evaluate our method qualitatively on this task.

\subsection{Implementation Details}

For all experiments, we use the same setting as NAFNet. The batch sizes are set to 8 and the training patches are 256 $\times$ 256 pixels. We use the Lion optimizer~\cite{chen2023symbolic} with $\beta_1=0.9$ and $\beta_2=0.99$. The initial learning rate is set to $3 \times 10^{-5}$ and decayed to 1e-7 by the Cosine scheduler. The noise level is fixed to 50 and the number of diffusion denoising steps is set to 100 for all tasks. We also augment the training data with random horizontal flips and 90 degree rotations. All models are implemented with PyTorch~\cite{paszke2019pytorch} and trained on a single A100 GPU for about 3 days.

For shadow removal and stereo super-resolution, we use the normal diffusion strategy and set the training iterations to 500\thinspace000. For HR dehazing and the Bokeh effect transformation, on the other hand, we incorporate the latent diffusion strategy. Specifically, we first train the U-Net on each dataset for 300\thinspace000 iterations, and then train the Refusion model based on the U-Net for 400\thinspace000 iterations. Here, all input patches are cropped to $1024 \times 1024$, while using U-Net to compress them to $128 \times 128$ pixels.

\begin{figure*}[t]
\begin{center}
\includegraphics[width=1.\linewidth]{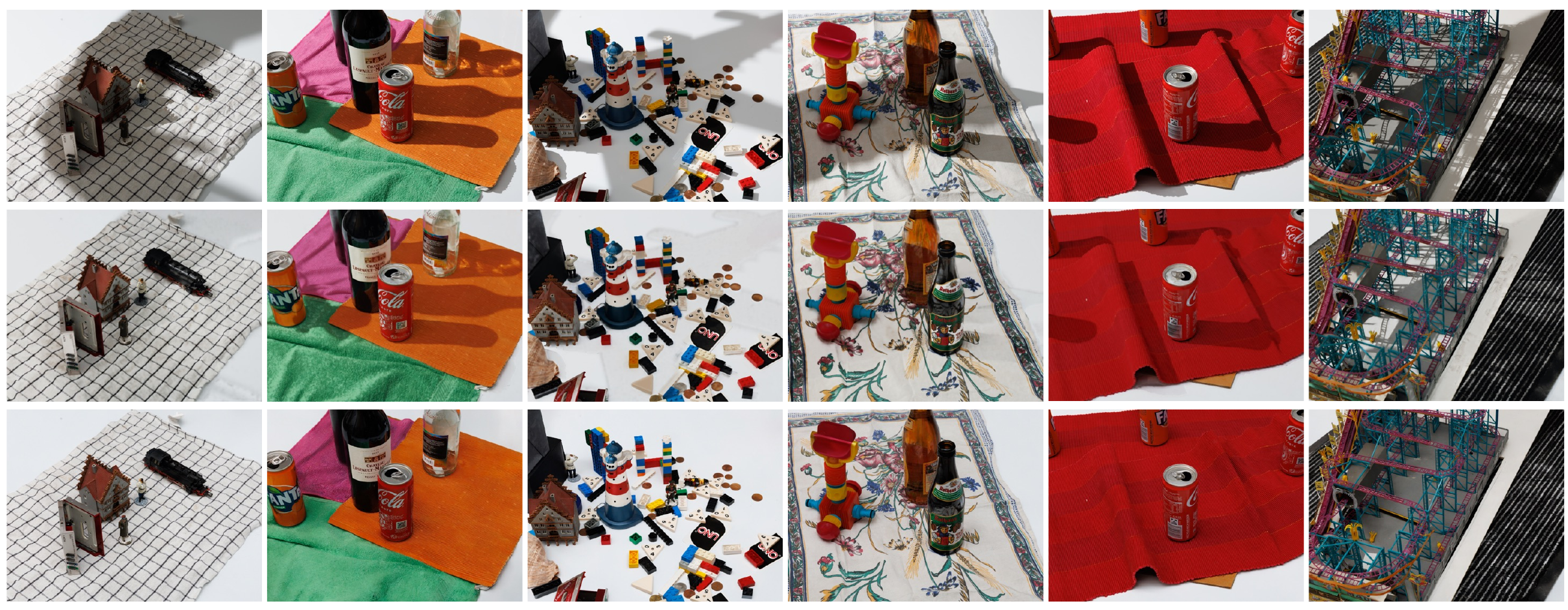}\vspace{-3.0mm}
\end{center}
    \caption{Visual results of our method and the U-Net baseline on the shadow removal task. Top row shows the input images and second row shows the results produced by the U-Net baseline. Bottom row shows the shadow-free results generated by our method.}
\label{fig:result_shadow}
\end{figure*}

\begin{table}[t]
\renewcommand{\arraystretch}{1.2}
\caption{Comparison of the proposed Refusion with IR-SDE~\cite{luo2023image} and NAFSSR~\cite{chu2022nafssr} on the stereo super-resolution validation dataset. The proposed Refusion achieves significantly better performance than IR-SDE across all metrics, and outperforms NAFSSR in terms of perceptual scores.}
\label{table:ssr}
\centering

\resizebox{1.\linewidth}{!}{
\begin{tabular}{lccccc}
\toprule
Method    & PSNR$\uparrow$       & SSIM$\uparrow$       & LPIPS$\downarrow$     & FID$\downarrow$   & Runtime$\downarrow$   \\ \midrule
NAFSSR~\cite{chu2022nafssr}	    & 23.81	    & 0.7247	    & 0.335    & 34.86   & 5.2s   \\ 
IR-SDE~\cite{luo2023image}	& 20.34	& 0.5841	    & 0.197  & 25.57  & 91.3s \\
Refusion	& 21.21	& 0.6336	    & \textbf{0.155}  & \textbf{22.43}  & 64.1s  \\
\bottomrule
\end{tabular}}
\end{table}

\begin{table}[t]
\renewcommand{\arraystretch}{1.2}
\caption{Comparison of the proposed Refusion with IR-SDE~\cite{luo2023image}, DHAN~\cite{cun2020towards} and an L1 loss trained U-Net baseline on the shadow removal dataset. Our proposed Refusion achieves the best restoration performance overall.}
\label{table:shadow}
\centering
\resizebox{1.\linewidth}{!}{
\begin{tabular}{lcccccc}
\toprule
Method    & PSNR$\uparrow$       & SSIM$\uparrow$       & RMSE$\downarrow$     & LPIPS$\downarrow$   & FID$\downarrow$   & Runtime$\downarrow$  \\ \midrule
DHAN~\cite{cun2020towards}	    & 20.42	    & 0.6986	   & 24.29   & 0.247  & 109.35   & 0.4s  \\
IR-SDE~\cite{luo2023image}	    & 20.30	    & 0.6639	  & 24.63     & 0.152  & 74.35   & 175.8s  \\
U-Net baseline	    & 20.69	    & 0.7172	& 23.55    & 0.236  & 102.1  & 1.62s  \\ 
Refusion	& 21.88	   & 0.6977	   & 20.53    & \textbf{0.121}  & \textbf{56.22}  & 38.4s  \\
\bottomrule
\end{tabular}}
\end{table}

\begin{table}[t]
\renewcommand{\arraystretch}{1.2}
\caption{Comparison of our methods with Restormer~\cite{zamir2022restormer} on the Bokeh Effect Transformation dataset. Our Refusion with latent strategy can also achieve good performance.}
\label{table:bokeh}
\centering

\resizebox{1.\linewidth}{!}{
\begin{tabular}{lccccc}
\toprule
Method    & PSNR$\uparrow$       & SSIM$\uparrow$       & LPIPS$\downarrow$   & FID$\downarrow$   & Runtime$\downarrow$ \\ \midrule
Restormer~\cite{zamir2022restormer}	    & 41.12	    & 0.9779	    & 0.067  & 46.72  & 2.0s \\ 
Refusion	& 39.81	& 0.9615	    & 0.053  & \textbf{20.38}   & 36.0s   \\
Latent Refusion 	& 40.24	& 0.9721	    & \textbf{0.047} & 24.25  & 6.9s  \\
\bottomrule
\end{tabular}}
\end{table}

\begin{figure}[t]
\begin{center}
\includegraphics[width=.95\linewidth]{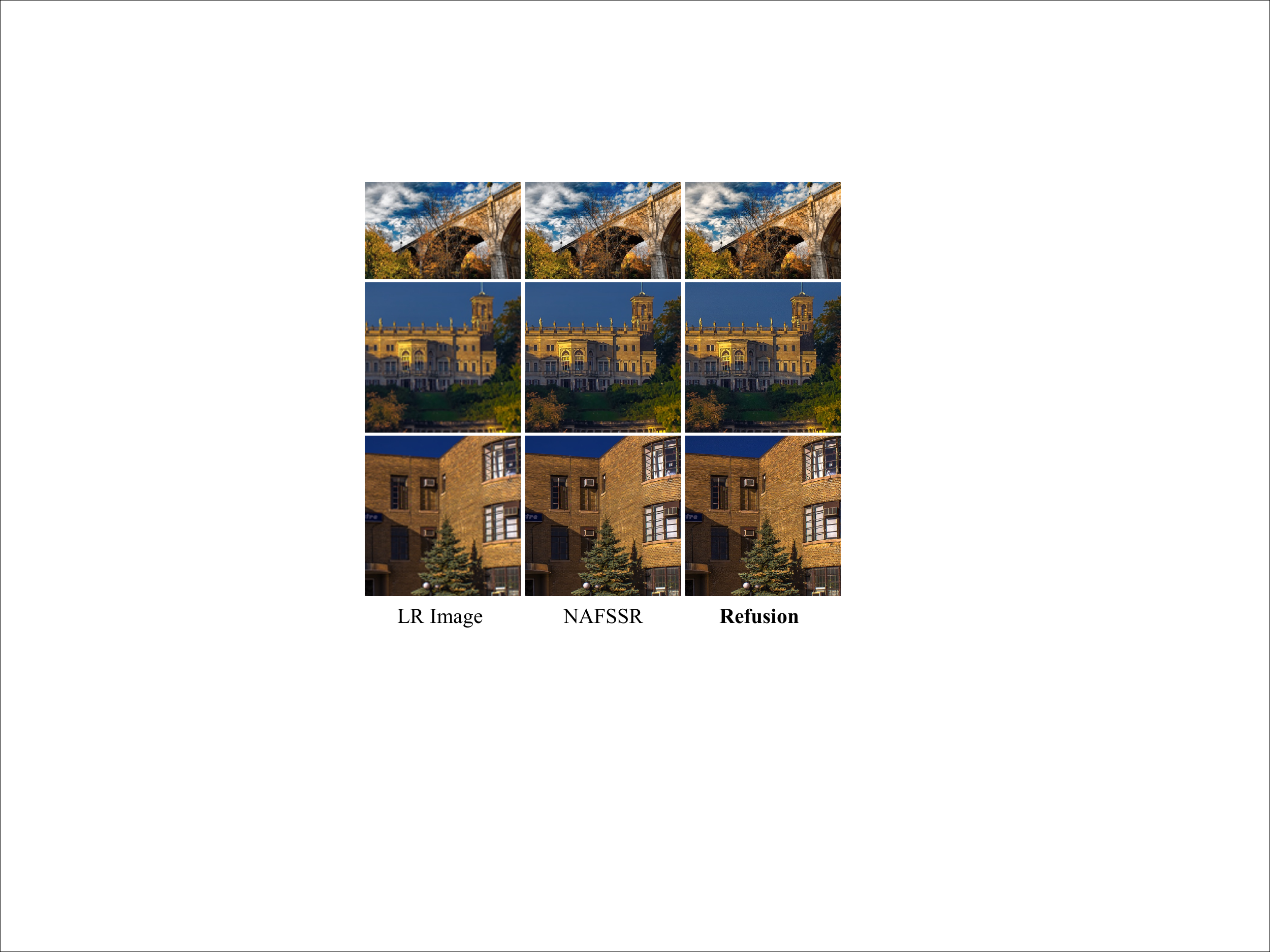}\vspace{-3.0mm}
\end{center}
    \caption{Visual results of our method and NAFSSR~\cite{chu2022nafssr} on the stereo super-resolution dataset.}
\label{fig:result_ssr}
\end{figure}

\begin{figure*}[ht]
\begin{center}
\includegraphics[width=1.\linewidth]{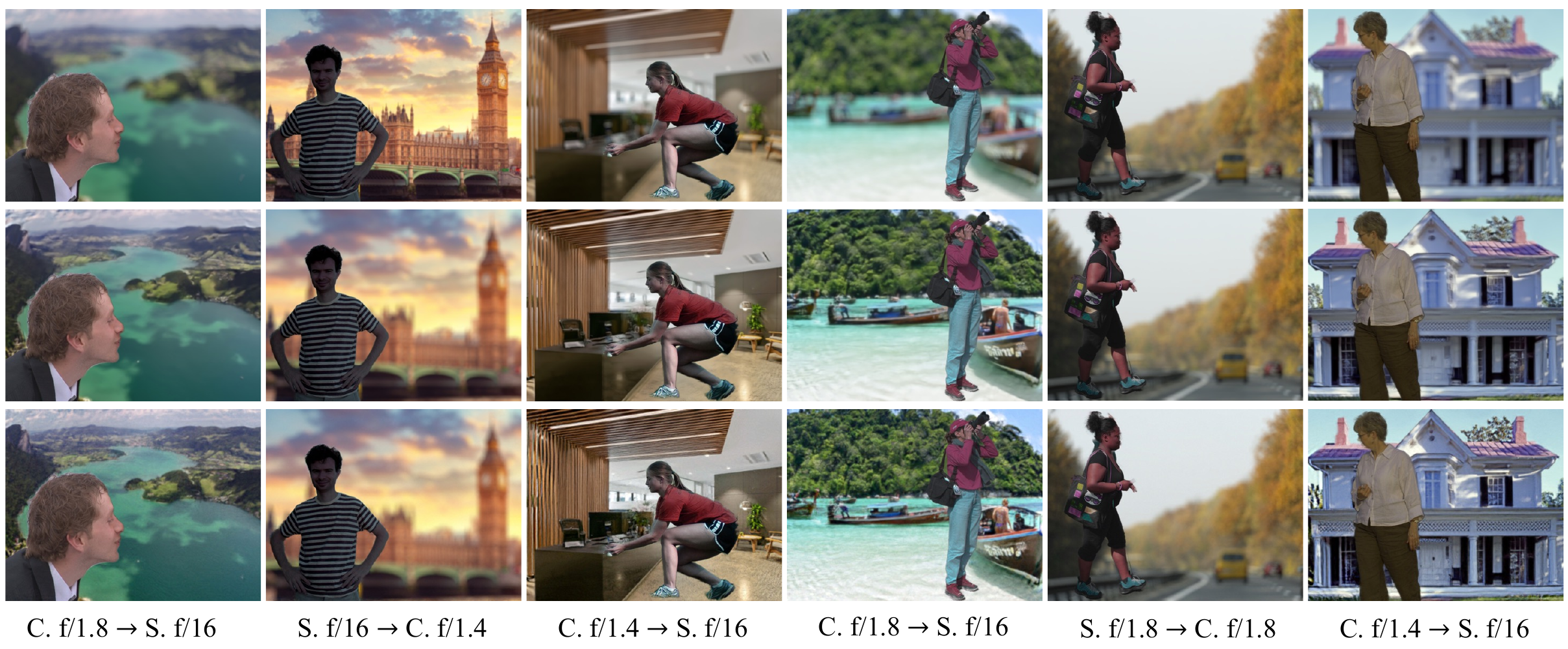}\vspace{-3.0mm}
\end{center}
    \caption{Visual results of our method and Restormer~\cite{zamir2022restormer} on the Bokeh effect transformation task. Top row shows the input images and second row shows the Restormer's results. Last row shows the transformed results generated by our method. In addition, lens transform information is shown in the bottom. `S. f/1.8 $\rightarrow$ C. f/16' means the image is transformed from \textit{Sony50mmf1.8BS} to \textit{Canon50mmf16.0BS}. }
\label{fig:result_bokeh}
\end{figure*}

\begin{figure*}[t]
\begin{center}
\includegraphics[width=1.\linewidth]{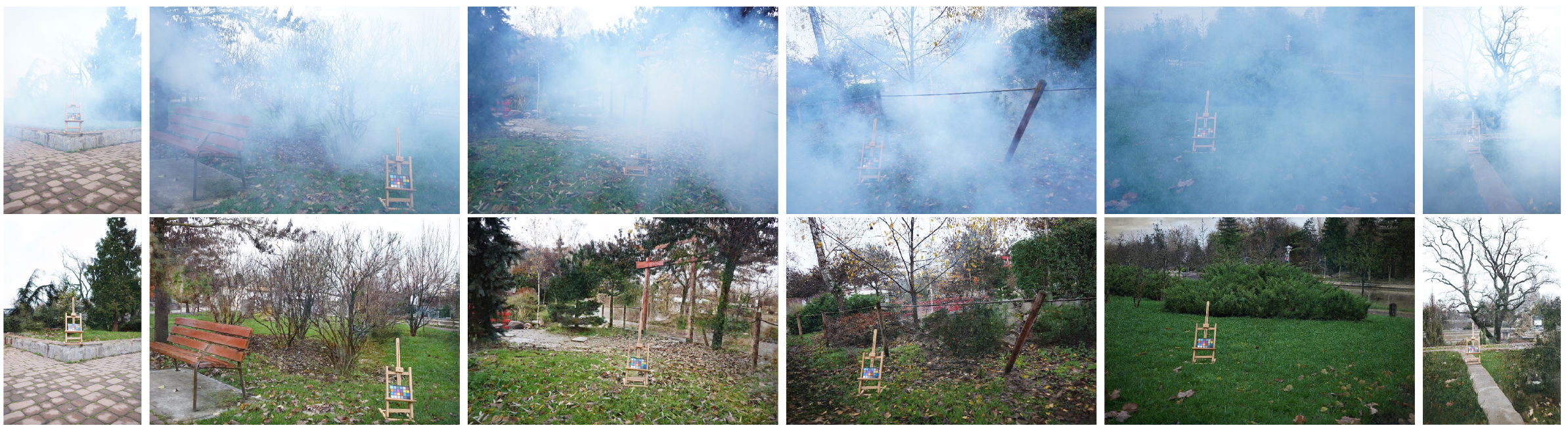}\vspace{-3.0mm}
\end{center}
    \caption{Visual results of our method on the HR Non-Homogeneous Dehazing task. Top and bottom row are inputs and outputs.}
\label{fig:result_haze}
\end{figure*}

\subsection{Experimental Results}

Since our Refusion method is proposed for realistic image restoration, we use the Learned Perceptual Image Patch Similarity (LPIPS)~\cite{zhang2018unreasonable} and Fr\'{e}chet inception distance (FID) score~\cite{heusel2017gans} as the main evaluation metrics, but we also report PSNR and SSIM for reference. For shadow removal, we further report the RMSE metric as previous approaches~\cite{cun2020towards,fu2021auto,chen2021canet}. Moreover, for each task, we also provide the runtime comparison to show the computational efficiency of our method against other baselines.

\vspace{5pt}
\noindent \textbf{Stereo Image Super-Resolution}. The quantitative comparison of our model with NAFSSR~\cite{chu2022nafssr} and IR-SDE~\cite{luo2023image} is shown in~\Cref{table:ssr}. NAFSSR achieves the best PSNR and SSIM scores, but performs inferior to IR-SDE and our Refusion in terms of LPIPS and FID. The proposed Refusion significantly improves the performance of IR-SDE across all metrics, demonstrating the effectiveness of our improved training strategies. Our method runs slower than NAFSSR, which directly predicts HQ images from LQ images, but clearly improves the runtime of our main baseline IR-SDE. The visual results are illustrated in~\Cref{fig:result_ssr}. Our Refusion produces sharper and clearer images than NAFSSR.

\vspace{5pt}
\noindent \textbf{Image Shadow Removal}. For this task, we compare our method with IR-SDE and a U-Net baseline model which uses the same network architecture as IR-SDE but is trained to directly predict HQ images via an $L_1$ loss. We also compare our method with DHAN~\cite{cun2020towards}, a well-established shadow removal model. The quantitative results are shown in~\Cref{table:shadow}. The proposed Refusion clearly achieves the best restoration performance overall. Moreover, Refusion runs significantly faster than IR-SDE. In the experiment, we find that all training image pairs have slight position shifts and that the luminance of each input and ground truth image is different, which may cause the $L_1$ loss trained model to learn the shift and luminance rather than shadow removal. Thus its performance is lower than our Refusion even in terms of PSNR. The qualitative comparison in~\Cref{fig:result_shadow} also demonstrates superior performance of Refusion.

\begin{table}[t]
\renewcommand{\arraystretch}{1.2}
\caption{Comparison of the parameters and flops. Flops are calculated on an image with size $256 \times 256$. Note that the additional U-Net for latent Refusion also has 6M parameters and 70G Flops, but it will be only run once for an image at test time.}
\label{table:param}
\centering
\resizebox{.85\linewidth}{!}{
\begin{tabular}{lccc}
\toprule
Method    & IR-SDE~\cite{luo2023image}       & Refusion       & Latent Refusion     \\ \midrule
\#Params	    & 135.3M	    & 131.4 M	    & 131.4M  \\ 
Flops	& 119.1G	& 63.4G	    & 4.0G   \\
\bottomrule
\end{tabular}
}
\end{table}

\vspace{5pt}
\noindent \textbf{Bokeh Effect Transformation}. We apply our latent Refusion model to this task, with an image downscale factor set to 4. As shown in~\Cref{table:bokeh}, both of our methods achieve better perceptual performance than Restormer~\cite{zamir2022restormer}. The latent Refusion achieves a better LPIPS score but slightly worse FID than Refusion. It also runs much faster than Refusion, getting close to the runtime of Restormer. The visual comparison is shown in~\Cref{fig:result_bokeh}. As one can see, our method produces sharper results than Restormer when transforming the bokeh effect from blurry to clear. We also provide a comparison of model complexities in~\Cref{table:param}. As can be observed, the latent strategy reduces computation flops about 15$\times$ compared to the original Refusion model, which significantly improves the applicability.

\vspace{5pt}
\noindent \textbf{HR Non-Homogeneous Dehazing}. The visual results of dehazing are shown in~\Cref{fig:result_haze}. As can be observed, most fog is successfully removed by our latent Refusion model. Note that all images in the HR dehazing dataset have $6000 \times 4000 \times 3$ pixels. With such large image sizes, IR-SDE and Refusion are not even able to process a complete image at test time, and other diffusion models which use additional self-attention mechanisms would be even more computationally expensive. By performing the restoration in a low-resolution latent space, our latent Refusion model can be applied also in this highly challenging setting.

\subsection{NTIRE 2023 Challenge Results}

To further validate the proposed method, we participated in 4 NTIRE 2023 Challenges~\cite{ancuti2023ntire,vasseiz2023ntireshr,conde2023ntire_bokeh,Wang2023NTIRE} corresponding to the aforementioned tasks. Our Refusion reached the final ranking of all competitions and even won the 2$^{\text{nd}}$ place in Shadow Removal task~\cite{vasseiz2023ntireshr}. The top ranked teams and their results in shadow removal are shown in~\Cref{table:ranking}. It is worth noting that \textbf{our method achieves the highest perceptual scores} in terms of LPIPS and \textit{Mean Opinion Score} (MOS).

\begin{table}[t]
\renewcommand{\arraystretch}{1.}
\caption{Final ranking of the Shadow Removal Challenge~\cite{vasseiz2023ntireshr}.}
\label{table:ranking}
\centering

\resizebox{.95\linewidth}{!}{
\begin{tabular}{cccccc}
\toprule
Rank  & Team  & PSNR$\uparrow$       & SSIM$\uparrow$       & LPIPS$\downarrow$   & MOS$\uparrow$   \\ \midrule
1	 & MTCV   & \textbf{22.36}	    & \textbf{0.7}	    & 0.182  & 8.31  \\ 
2	 & \textbf{IR-SDE (Ours)}   & 19.6	    & 0.58	    & \textbf{0.149}  & \textbf{8.94}  \\ 
3	 & SRDM   & 22.2	    & 0.69	    & 0.269  & 7.33  \\ 
4	 & SYU-HnVLab   & 21.25	    & 0.67	    & 0.217  & 6.84  \\ 
5	 & MegSRD   & 17.36	    & 0.53	    & 0.198  & 8.81  \\ 
6	 & UM-JTG   & 21.7	    & 0.69	    & 0.283  & 6.81  \\ 
\bottomrule
\end{tabular}}
\end{table}

%% file: sections/conclusion.tex
\section{Conclusion}

In this paper, we present several techniques to improve the applicability of diffusion-based image restoration. The resulting model, named \textit{Refusion}, is successfully applied to various image restoration tasks and it achieves the best perceptual performance in the NTIRE 2023 Shadow Removal Challenge. To process large-size images, we further propose a U-Net based latent Refusion model that compresses the input image to a low-resolution latent space in which it performs diffusion to improve the model efficiency. Since input image information is also captured in the hidden vectors connected to the decoder, we are able to recover images with more accurate details. Our latent Refusion model can even run on images of size $6000 \times 4000 \times 3$ pixels.